\documentclass[10pt,twocolumn,letterpaper]{article}

\usepackage[pagenumbers]{wacv} 

\usepackage{tabularx}
\usepackage{enumitem}
\usepackage{colortbl}
\usepackage{xcolor}
\usepackage{multirow}
\usepackage{booktabs}
\usepackage{amsmath}
\usepackage{algorithm}
\usepackage{algorithmic}
\usepackage{amssymb}
\usepackage{float}
\usepackage{pifont}
\usepackage{graphicx}
\usepackage{microtype}

\providecommand{\eg}{\emph{e.g.}}

\definecolor{avggray}{gray}{0.88}

\usepackage{xr-hyper}
\definecolor{wacvblue}{rgb}{0.21,0.49,0.74}
\usepackage[pagebackref,breaklinks,colorlinks,allcolors=wacvblue]{hyperref}

\title{To Memories and Beyond: From Remembering to Knowing You across Long-Term Multimodal Personal Archives}

\author{%
  Wenqi Zhou$^{1,2}$ 
  \quad Zhuorui Yu$^{2}$ 
  \quad Kaiao Wen$^1$ 
  \quad Hao Zheng$^{1,2}$ 
  \quad Xinyi Zheng$^1$\\
  \quad Peiran Wu$^{1,2}$ 
  \quad Enmin Zhou$^{2}$
  \quad Chi-Hao Wu$^{2\dagger}$
  \quad Junxiao Shen$^{1,2\dagger}$
  \\\\
  \textsuperscript{1}University of Bristol
  \quad \textsuperscript{2}Memories.ai Research
}
\begin{document}
\maketitle
\footnotetext[2]{Corresponding author.}
\begin{abstract}
As AI systems evolve into personalized digital companions, a central capability is reasoning over a user's long-term personal history: not merely storing past events, but tracking longitudinal experiences and evolving preferences. Progress here is bottlenecked by evaluation, existing long-term memory benchmarks are largely synthetic and text-only, they overlook the visual records that anchor everyday human memory, lack the authentic and causally connected longitudinal data that real personalization demands, and consequently remain confined to shallow factual recall. We introduce ReaLMem (Real-world Long-term Multimodal Memory), the first benchmark built from authentic multi-year personal visual archives, paired with first-person subjective annotations. ReaLMem evaluates models across three cognitive tiers of increasing difficulty: factual recall, persona inference, and predictive personalization. We further propose ChronoProfiler, a temporal-weighting profiling module that computes temporal stability scores for user attributes and applies them as a salience prior, resolving conflicts among temporally inconsistent preferences and helping models compound multiple co-active preferences in complex personalized decisions. Extensive evaluation of frontier multimodal large language models (MLLMs) and memory systems on ReaLMem reveals predictive personalization as a consistent ceiling, exposes clear performance gaps and bottlenecks between MLLMs and memory systems, and shows that high-quality, temporally informed representations substantially improve personalization. Together, ReaLMem and ChronoProfiler provide an authentic testbed and a simple, effective mechanism for long-term personalization, laying a foundation for future research on lifelong AI companions.
\end{abstract}

\section{Introduction}
As AI evolves toward Artificial General Intelligence (AGI), human--computer interaction is shifting from one-off instruction execution toward persistent, personalized digital companions. Long-term memory is the cornerstone of this shift \cite{lee2024towards, zhang2025survey}: it lets a model accumulate, track, and reason over a user's experiences and evolving preferences across years, moving beyond isolated factual Q\&A toward responses tailored to an individual's history. Accordingly, recent work has begun to build benchmarks for long-context, long-horizon personal memory \cite{wu2024longmemeval, maharana2024evaluating, tavakoli2025beyond}. Yet progress toward genuinely personalized long-term assistants is held back by two limitations of current evaluation---the data and the tasks.

\begin{figure*}[ht!]
  \centering
  \includegraphics[width=0.95\linewidth]{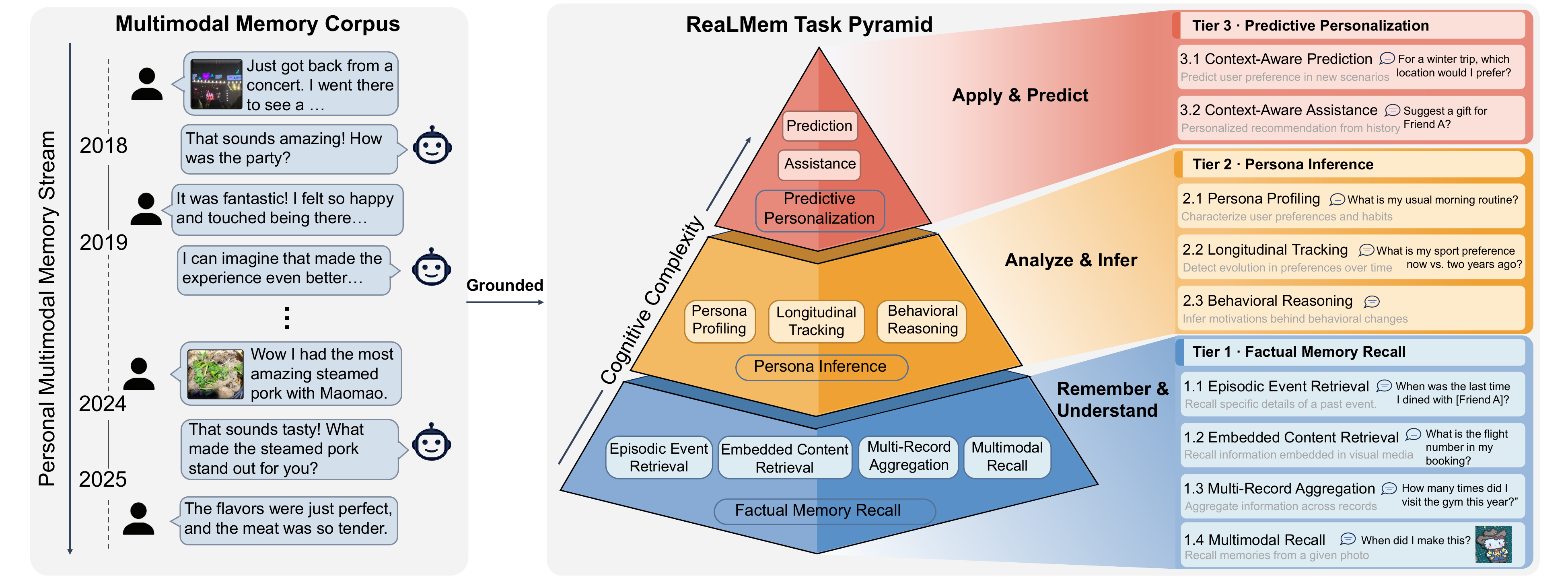}
  \caption{Overview of ReaLMem: (left) example personal multimodal memory stream over years; (right) the three-tier task pyramid grounded in this memory, rising in cognitive complexity from Factual Memory Recall (Tier~1) and Persona Inference (Tier~2) to Predictive Personalization (Tier~3), with each of the nine subtasks shown alongside its definition and an example query.}
  \label{fig:task_tax}
\end{figure*}

First, current benchmarks are largely inauthentic and vision-blind. Constrained by privacy and cost, they are predominantly built top-down from text-only data synthesized by large language models (LLMs) \cite{zhong2024memorybank, du2024perltqa, jiang2025know, jiang2512personamem}, or stretched to long context by mechanically inserting irrelevant passages (\eg, ``needle-in-a-haystack'' padding) \cite{hsieh2024ruler, schuster2025needle}. Such synthesis disrupts the natural temporal and causal structure of memory and rarely captures the messiness of genuine human experience. More fundamentally, it overlooks the visual record---the photos and videos that anchor much of everyday human memory~\cite{Brady2008VisualLM}---leaving evaluation disconnected from the physical world.

Second, as a direct consequence, evaluation stops at shallow factual recall. Because synthetic corpora lack the authentic, causally connected longitudinal data that accrues in real life, they cannot support the deeper competencies that genuine personalization demands. Human preferences are dynamic and context-dependent, shaped by the interplay of situation and emotion; assessing personalized assistants therefore requires moving beyond retrieval toward understanding how preferences evolve, reasoning about why they change, and predicting future choices.

These limitations raise two questions: (a) what data can ground the evaluation of authentic long-term personalization; (b) how should tasks be designed to comprehensively evaluate the ability to understand human personalization?

Egocentric and lifelog vision~\cite{Grauman2021Ego4DAT,Dong2025EgoLifeTE,Damen2020RescalingEV} has begun to study personal visual data directly, but mainly for short-horizon perception and retrieval, and typically without the data owner's first-person subjective annotations---two gaps that have kept personalization underdeveloped. We argue that the personal media archives already stored on people's devices---especially photos and videos---are instead a natural and powerful substrate for \emph{long-term} multimodal memory: spanning years, they record where users went, what they did, and what mattered to them, preserving the visual and contextual anchors that text-only logs discard. Treated as long-term memory and paired with the owner's own perspective, these archives open a path to assistants that provide proactive, context-aware, and personalized help grounded in visual memory.

Building on above insights, we depart from the synthetic-data paradigm and introduce ReaLMem, the first multimodal long-term memory benchmark dataset derived from authentic personal visual archives paired with first-person subjective annotations. To structure evaluation, and inspired by Bloom's taxonomy of educational objectives \cite{bloom1956handbook}, we organize tasks into three progressively harder cognitive tiers (Fig.~\ref{fig:task_tax}): from factual recall, through persona inference, to predictive personalization over a user's history.

Realistic personalization at this scale also faces a concrete architectural obstacle: feeding ultra-long multimodal histories to a model is computationally costly and prone to the lost-in-the-middle effect \cite{liu2024lost}, while existing approaches flatten historical preferences \cite{sun2025preference, westhausser2025enabling}, treating one-off mentions and long-held habits alike and leaving no principled way to arbitrate when preferences conflict or compete. As a targeted remedy, we propose ChronoProfiler, a plug-and-play user-profiling module that incrementally distills a structured persona from longitudinal history and assigns each item a temporal stability score. Used as a salience prior, this score resolves temporal conflicts and helps the model weigh and compound co-active preferences in complex personalized decisions.

In summary, our contributions are:

\noindent{\small$\bullet$} We introduce ReaLMem, the first multimodal long-term memory benchmark built from authentic personal visual archives and first-person subjective annotations, comprising 2,508 multimodal sessions spanning more than six years with genuine spatio-temporal metadata.

\noindent{\small$\bullet$} We design a three-tier cognitive evaluation hierarchy---Factual Memory Recall, Persona Inference, and Predictive Personalization---instantiated by 1,629 QA pairs for fine-grained assessment of personalization.

\noindent{\small$\bullet$} We propose ChronoProfiler, a temporal-aware user-profiling module that turns a temporal stability score into a salience prior for preference-weighted personalized tasks.

\noindent{\small$\bullet$} We comprehensively evaluate multimodal LLMs (MLLMs) and memory systems on ReaLMem, revealing performance gaps, modality dependencies, and open challenges.

\section{Related Works}
\subsection{Personal Memory Benchmarks}
Recent personal memory benchmarks \cite{zhong2024memorybank, xu2022long, jang-etal-2023-conversation, li2025hello} approach the long-term memory challenge along three complementary trajectories. A first line broadens modality and source coverage by generating multi-session, multimodal dialogues from causal temporal event graphs \cite{maharana2024evaluating}. A second line embeds user preferences implicitly within conversational context, testing deductive reasoning over fragmented and incidental signals \cite{du2024perltqa, jiang2025know, jiang2512personamem}. A third line scales context length via a ``needle-in-a-haystack'' construction that injects critical evidence into vast amounts of distractor dialogues~\cite{wu2024longmemeval}, reaching contexts of up to 1.5\,M tokens.

Despite their progress, these benchmarks share three structural limitations rooted in their synthetic origin. (i) Modality narrowness, or vision-blindness: most evaluation data is text-only, leaving the visual signals that anchor everyday human memory under-explored. (ii) Density loss and structural flattening: persona templates and temporal graphs, when realized through LLM synthesis, erase the modality cues, causal continuity, and natural temporal alignment of lived experience, producing dialogues that lack genuine contextual nuance and unpredictability. (iii) Limited cognitive depth: as a consequence, evaluation is largely confined to factual retrieval, with little room for the deeper reasoning required for multi-preference compounding or dynamic conflict resolution under real-world constraints.

To close this gap, we ground evaluation in authentic long-term multimodal personal data rather than synthetic substrates. Building on real, longitudinal records of participants' daily lives---where every moment carries its own implicit personal context---we introduce the first benchmark constructed from genuine multimodal personal histories, together with a three-tier task hierarchy that rigorously evaluates factual recall, preference and behavior inference, and complex predictive personalization.

\subsection{Egocentric and Lifelog Vision}
Some research works study personal visual data directly. Egocentric video understanding builds large first-person datasets for activity and interaction recognition~\cite{Grauman2021Ego4DAT, Damen2020RescalingEV, Grauman2023EgoExo4DUS}, and lifelog retrieval organizes and searches wearable-camera or personal-photo streams~\cite{Gurrin2016OverviewON, Gurrin2014LifeLoggingPB}. Recent efforts push toward longer horizons and assistant-style use---extremely long egocentric video understanding~\cite{Zhou2025XLeBenchAB} and egocentric life assistants~\cite{Dong2025EgoLifeTE}---while OmniQuery~\cite{li2025omniquery}, closest to our setting, answers personal questions over captured multimodal memories. 

Yet these efforts largely operate on continuous egocentric video over hours to days, rely on third-person or task labels rather than the data owner's own subjective annotations, and stop short of the higher-order persona inference and predictive personalization our tasks require. ReaLMem advances this line from perception toward long-term personalized cognition, pairing multi-year personal visual archives with first-person annotations and a graded, cognitively tiered evaluation.

\begin{figure*}[ht]
  \centering
  \includegraphics[width=\linewidth]{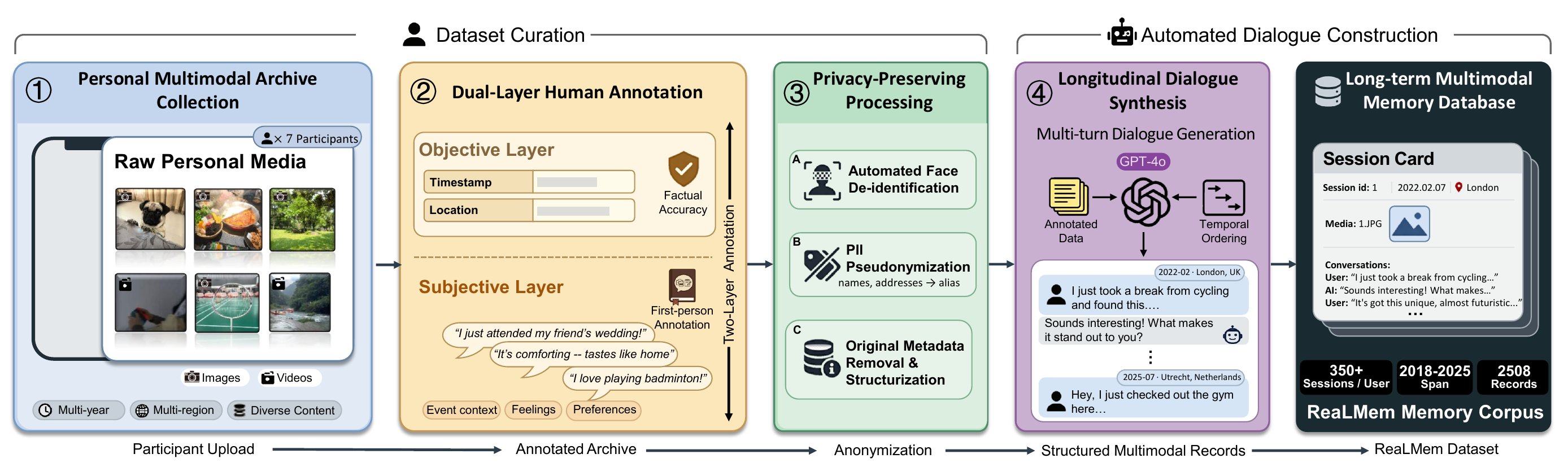}
  \caption{The ReaLMem construction pipeline: raw personal photos and videos are annotated by their owners (objective and subjective layers), anonymized, and woven into chronological multimodal sessions.}
  \label{fig:data_construct}
\end{figure*}
\subsection{Memory Mechanism for LLM}
Endowing LLMs with long-term memory has been pursued along two complementary directions. The first scales raw access to history through ever-longer context windows or retrieval-augmented generation (RAG)~\cite{lewis2020retrieval, wang2023augmenting, xiao2024efficient}, which fetches semantically similar historical snippets at query time. The second introduces structured memory architectures that organize personal history into reusable representations, including knowledge-graph memories~\cite{chhikara2025mem0,rasmussen2501zep}, hierarchical memory trees~\cite{li2025toward}, and OS-style memory frameworks that manage memory as a lifecycle-controlled resource~\cite{li2505memos,hu2026evermemos}.

Despite these advances, existing systems share two limitations directly relevant to long-horizon personalization. (i) Preference flattening: retrieval and storage treat all historical signals as equally valid, providing no principled mechanism to differentiate long-held habits from one-off mentions, so the model is left to arbitrate competing preferences on its own. (ii) Weak temporal awareness: timestamps, when present, are typically used only as retrieval filters rather than as first-class priors over content, leading to unresolved conflicts between outdated and recent records during reasoning. Together, these gaps limit reliable behavior on tasks that require synthesizing heterogeneous, time-varying signals into a coherent personalized decision.

In response, ChronoProfiler builds a per-user profile in which every item is grounded by timestamped evidence and weighted by a temporal stability score derived from its temporal features and used as a salience prior, enabling memory systems to resolve temporal conflicts and prioritize stable preferences without modifying the underlying LLM.

\section{ReaLMem}

We introduce ReaLMem (Real-world Long-term Multimodal Memory). Unlike prior benchmarks built from pre-defined, synthesized personas, ReaLMem is constructed from authentic personal visual archives---photos and videos spanning years of real life---paired with first-person subjective annotations from the data owners themselves. This grounding in real multimodal experience and the owner's own perspective provides a rigorous basis for studying memory retention, preference evolution, and personalized assistance. We describe the dataset construction and task design below.

\subsection{Dataset Construction}

We construct ReaLMem through a participant-centric pipeline that turns raw personal visual archives into structured, high-fidelity multimodal memory sequences (Fig.~\ref{fig:data_construct}). We recruited 7 participants with dense, multi-year media collections, each contributing over 350 photos and videos that capture meaningful events, daily routines, and preference-revealing behaviors (\eg, dietary habits, long-term hobbies), all grounded with reliable spatio-temporal metadata. We deliberately favor depth over breadth: collecting truly authentic, long-horizon archives is bounded by acquisition cost and strict privacy requirements, so a small set of richly annotated, real participants offers grounding that large synthetic or crowdsourced corpora cannot. To ensure both factual accuracy and subjective authenticity, each participant followed a dual-layer annotation protocol: an objective tier verified structured metadata, while a subjective tier recorded each record's first-person context---the personal event context, the owner's feelings, and implicit preference signals. All data collection, processing, and annotation followed an ethics protocol approved by our organization's ethical review board, with written informed consent from every participant, details of ethical consideration are provided in Appendix~\ref{app:ethics}.

After stringent anonymization: automated face blurring with RetinaFace~\cite{Deng2020CVPR}, personally identifiable information (PII) pseudonymization, and metadata reorganization (details in Appendix~\ref{app:anonymization}), we use GPT-4o~\cite{Hurst2024GPT4oSC} to weave each participant's records and annotations, in chronological order, into natural multi-turn dialogues. Crucially, the model only renders the narrative form: every fact, timestamp, and preference originates from the real records and the owner's annotations, not from model invention. As summarized in Tab.~\ref{tab:data_stats}, the resulting data is diverse across time, space, and semantics: it spans 2018--2025, covers more than 40 countries and regions and ten everyday scenario categories, and comprises 2,333 images and 175 videos. In total, ReaLMem contains 2,508 multimodal sessions (about 358 per participant), forming long-term and coherent personal trajectories.

\begin{table}[h]
\centering
\caption{Statistics of the ReaLMem dataset. $T$ denotes the year of
  data collection.
  Country counts reflect the top-5 capturing locations by media volume.}
\label{tab:data_stats}
\small
\setlength{\tabcolsep}{4pt}
\renewcommand{\arraystretch}{0.9}
\begin{tabular}{@{}lrlr@{}}
\toprule
\multicolumn{2}{c}{\textbf{Year (rel.\ to $T$)}} & \multicolumn{2}{c}{\textbf{Media Type}} \\
\cmidrule(r){1-2}\cmidrule(l){3-4}
$T$              & 857 & Image  & 2,333 \\
$T{-}1$          & 799 & Video  &   175 \\
$T{-}2$          & 327 & \multicolumn{2}{c}{\textbf{Country (Top 5)}} \\
\cmidrule(l){3-4}
$T{-}3$          & 285 & China          & 941 \\
$T{-}4$          & 148 & United Kingdom & 842 \\
$T{-}5$          &  58 & Japan          & 127 \\
${\leq}\,T{-}6$  &  34 & Spain          & 114 \\
                 &     & France         &  74 \\
\midrule
\multicolumn{4}{c}{\textbf{Scenario Type}} \\
\cmidrule{1-4}
Leisure \& Entertain.  & 815 & Sports \& Physical Act. & 184 \\
Nature \& Outdoor      & 790 & Special Events          & 175 \\
Home \& Routine        & 400 & Travel \& Mobility      & 145 \\
Social Interaction     & 399 & Work \& Study           & 128 \\
Fashion                &  27 & Medical \& Care         &  8 \\
\bottomrule
\end{tabular}
\end{table}

\subsection{Task Design}
Inspired by cognitive theories \cite{bloom1956handbook, tulving1972episodic} of human memory and reasoning, we design a hierarchical benchmark that reflects the progressive nature of human memory and cognition, framing personalized AI capability as a transition from information access to behavioral understanding and decision-oriented application.
Specifically, we define three categories of increasing complexity, with per-subtask definitions and example queries shown in Fig.~\ref{fig:task_tax} (the full taxonomy with per-subtask QA counts is given in Tab.~\ref{tab:task_overview}). \emph{Factual Memory Recall} (T1 task) tests foundational memory access grounded in evidence---retrieving event details (task~1.1), aggregating or comparing across records (task~1.3), and aligning information across the long-term archive, including content available only visually, such as embedded text in visual data (task~1.2) and memories grounded in a given photo (task~1.4). \emph{Persona Inference} (T2 task) moves from explicit observation to implicit patterns: profiling stable habits (task~2.1), tracking how preferences evolve over time (task~2.2), and reasoning about the motivations behind behavioral change (task~2.3). \emph{Predictive Personalization} (T3 task) turns this understanding into application: predicting users' choices in novel scenarios (task~3.1) and generating personalized recommendations based on personal history (task~3.2). Unlike prior benchmarks based on synthetic personas or unimodal dialogue, this formulation is grounded in real multimodal experience and traces a continuous path from recall to decision-making, enabling fine-grained evaluation of long-term personalization.

\noindent\textbf{QA Instantiation.}
Each tier adopts a QA format aligned with its evaluation goal. T1 questions are open-ended, paired with a reference answer and explicit grounding evidence (data IDs). T2 questions augment the reference answer with a set of key supporting points that enable coverage evaluation of inherently subjective reasoning. T3 adopts a ranking paradigm in which each query presents multiple preference-constructed candidates, and the task is to recover the user's actual preference ordering. All 1{,}629 QA pairs are screened and refined by trained annotators. Additionally, T2 and T3 instances are verified by the original data contributors to ensure faithfulness to lived and subjective experience. Full construction procedures, quality-control criteria, and ranking-paradigm design are provided in Appendix~\ref{app:qa_generation}.

\section{ChronoProfiler}
\label{sec:chronoprofiler}
Existing memory systems face two failure modes: (i) Temporal Conflicts, where outdated and recent signals coexist in context (\eg, ``vegan two years ago'' vs.\ ``BBQ last week''), confusing the model about the user's current state; and (ii) Intra-class Competition, where multiple same-category preferences (\eg, hotpot vs.\ sushi) compete without a principled arbitration mechanism. Both issues are especially damaging on tasks that require understanding and reasoning over long-term personal information, where the model must synthesize heterogeneous, time-varying signals into a coherent judgment. 
Inspired by memory consolidation~\cite{klinzing2019mechanisms, mcclelland1995there}, in which repeatedly activated and recent experiences are reinforced while transient ones decay, we propose ChronoProfiler, which incrementally builds an increasingly comprehensive user profile as the history accumulates and assigns each user attribute item a temporal stability (TS) score that serves as a salience prior at inference. The module runs in two stages, detailed below: 

\textbf{(1) Incremental Profile Construction.}
We process the dialogue session-by-session, prompting an LLM to extract structured attributes into five semantic categories (Demographics, Preferences, Behavioral Habits, Hobbies, Meaningful Facts). Each candidate is semantically compared against existing entries in its category and resolved via one of four operations: \textsc{New}, \textsc{Duplicate}, \textsc{Update}, or \textsc{Conflict}, yielding a comprehensive profile in which every item carries a set of timestamped evidence. Implementation details are provided in Appendix~\ref{app:chronoprofiler}.

\textbf{(2) Temporal Stability Scoring.}
For each profile item $p$, let $\mathcal{S}_p$ denote the set of timestamps at which $p$ has been mentioned across the dialogue history, and let $t_{\mathrm{first}}=\min\mathcal{S}_p$ and $t_{\mathrm{last}}=\max\mathcal{S}_p$ be its earliest and latest mentions. Given a reference time $T_{\mathrm{ref}}$ (set to the query timestamp at inference), we summarize the lifecycle of $p$ with three features:
\begin{equation*}
\begin{aligned}
s_{\mathrm{sup}} &= 1 - e^{-|\mathcal{S}_p| / \tau_s}, \\
s_{\mathrm{dur}} &= 1 - e^{-(t_{\mathrm{last}} - t_{\mathrm{first}}) / \tau_d}, \\
s_{\mathrm{rec}} &= e^{-(T_{\mathrm{ref}} - t_{\mathrm{last}}) / \tau_r}.
\end{aligned}
\end{equation*}
The Support score $s_{\mathrm{sup}}$ measures \emph{how often} $p$ is mentioned and saturates as $|\mathcal{S}_p|$ exceeds the characteristic count $\tau_s$. The Duration score $s_{\mathrm{dur}}$ measures \emph{how long} the attribute persists, growing as the span $t_{\mathrm{last}}-t_{\mathrm{first}}$ exceeds the timescale $\tau_d$. The Recency score $s_{\mathrm{rec}}$ measures \emph{how recently} $p$ was reaffirmed, decaying exponentially in the elapsed time $T_{\mathrm{ref}}-t_{\mathrm{last}}$ with characteristic constant $\tau_r$.

Each feature's contribution adaptively per user via the entropy weight method~\cite{Shannon1948AMT, Zhu2020EffectivenessOE}, a label-free criterion-weighting technique from multi-criteria decision analysis. Intuitively, a lifecycle feature that is nearly constant across a user's profile items carries little discriminative information and should not drive prioritization, whereas a widely-varying feature is informative and is up-weighted. Let $\mathbf{X}\in\mathbb{R}^{m\times 3}$ collect the three feature scores $s_j(p)\in\{s_{\mathrm{sup}}, s_{\mathrm{dur}}, s_{\mathrm{rec}}\}$ over the user's $m$ profile items. For each feature column $j$ we normalize $p_{ij}=x_{ij}/\sum_i x_{ij}$, compute its Shannon entropy $e_j=-\tfrac{1}{\ln m}\sum_i p_{ij}\ln p_{ij}$, and weight it by the resulting information divergence:
\begin{equation*}
w_j = \frac{1-e_j}{\sum_{j'}\left(1-e_{j'}\right)}.
\end{equation*}
The raw stability of $p$ is the entropy-weighted sum $z_p = \sum_j w_j\, s_j(p)$. Finally, to obtain a salience prior comparable across users with heterogeneous engagement rhythms, we apply within-user min--max normalization:
\begin{equation*}
\mathrm{TS}(p) = \frac{z_p - \min_{q} z_q}{\max_{q} z_q - \min_{q} z_q},
\end{equation*}
spreading each user's items across $[0,1]$. Long-held and repeatedly reinforced preferences receive high TS (consolidated long-term memory), whereas recent but unaccumulated signals receive moderate TS (current salience without permanent commitment). The TS-scored profile is then injected into the LLM context as a salience prior, guiding the model to prioritize temporally stable signals during personalized decision-making. Parameter settings are provided in Appendix~\ref{app:chronoprofiler}.

\section{Experiments}
\begin{table*}[t]
\centering
\footnotesize
\setlength{\tabcolsep}{3.5pt}
\renewcommand{\arraystretch}{0.8}
\caption{%
  Main results on the \textsc{ReaLMem} benchmark (\%), all metrics are scaled to $[0,100]$.
  Subtask abbreviations: \textbf{Retr.}/\textbf{Embd.}/\textbf{Aggr.}/\textbf{M.R.} = Episodic Event Retrieval / Embedded Content Retrieval / Multi-Record Aggregation / Multimodal Recall; \textbf{Prof.}/\textbf{Track}/\textbf{Reas.} = Persona Profiling / Longitudinal Tracking / Behavioral Reasoning; \textbf{Pred.}/\textbf{Asst.} = Context-Aware Prediction / Context-Aware Assistance.
  Shaded \textbf{Avg.}\ columns report the task-level mean.
  Best per column \emph{within each block} in \textbf{bold}.%
}
\label{tab:main_results}
\resizebox{\textwidth}{!}{%
\begin{tabular}{@{}l c
  c c c c >{\columncolor{avggray}}c
  c c c >{\columncolor{avggray}}c
  c c >{\columncolor{avggray}}c@{}}
\toprule
\multirow{2}{*}{\textbf{Method}} & \multirow{2}{*}{\textbf{Avg. Token}}
  & \multicolumn{5}{c}{\textbf{T1}}
  & \multicolumn{4}{c}{\textbf{T2}}
  & \multicolumn{3}{c}{\textbf{T3}} \\
\cmidrule(lr){3-7}\cmidrule(lr){8-11}\cmidrule(lr){12-14}
  & & \textbf{Retr.} & \textbf{Embd.} & \textbf{Aggr.} & \textbf{M.R.} & \textbf{Avg.}
    & \textbf{Prof.} & \textbf{Track} & \textbf{Reas.} & \textbf{Avg.}
    & \textbf{Pred.} & \textbf{Asst.} & \textbf{Avg.} \\
\midrule
\multicolumn{14}{l}{\textit{Full Context}} \\[1pt]
\quad Gemini-3.0-Flash & 123k & \textbf{84.4} & \textbf{77.7} & \textbf{75.2} & \textbf{89.3} & \textbf{82.6} & 74.8 & \textbf{73.0} & \textbf{81.7} & \textbf{76.4} & \textbf{62.8} & 53.5 & \textbf{59.2} \\
\quad GPT-4.1-mini     & 123k & 78.0 & 70.4 & 43.8 & 65.5 & 72.6 & 74.2 & 67.4 & 81.0 & 74.3 & 60.4 & 54.0 & 57.9 \\
\quad GPT-5.4-mini     & 123k & 79.7 & 68.9 & 40.0 & 57.1 & 72.5 & \textbf{75.9} & 68.9 & 81.3 & 75.5 & 59.2 & \textbf{58.7} & 59.0 \\
\midrule
\multicolumn{14}{l}{\textit{Memory Systems (GPT-4.1-mini backbone)}} \\[1pt]
\quad Mem0             & 1.0k & 47.5 & 61.9 & 14.3 & 33.3 & 46.8 & 54.1 & 52.3 & 62.2 & 56.0 & 59.5 & \textbf{58.7} & \textbf{59.2} \\
\quad MemOS            & 1.1k & 57.3 & 57.7 & 30.5 & 40.5 & 54.0 & 59.8 & 52.5 & 63.5 & 58.8 & \textbf{61.3} & 54.0 & 58.4 \\
\quad EverMemOS        & 1.7k & \textbf{72.2} & \textbf{72.3} & \textbf{47.6} & \textbf{60.7} & \textbf{69.4} & \textbf{72.8} & \textbf{64.9} & \textbf{77.6} & \textbf{72.0} & 57.4 & 55.9 & 56.8 \\
\bottomrule
\end{tabular}}
\end{table*}
\subsection{Experimental Setup}
We benchmark three categories of frontier approaches on \textsc{ReaLMem}: (1)~\textbf{Full-context MLLMs}---Gemini-3.0-Flash~\cite{gemini-3.0-flash}, GPT-4.1-mini~\cite{gpt-4.1-mini}, and GPT-5.4-mini~\cite{gpt-5.4-mini}---receive the complete captioned history as a practical upper bound under unrestricted access. (2)~\textbf{Memory systems}---Mem0~\cite{chhikara2025mem0}, MemOS~\cite{li2505memos}, and EverMemOS~\cite{hu2026evermemos}---compress per-user histories into persistent structured memories and retrieve relevant content at query time; to isolate the memory architecture from the underlying LLM, all memory systems share a unified GPT-4.1-mini answering backbone. (3)~\textbf{ChronoProfiler (Ours)} performs a controlled \emph{profile swap}: holding EverMemOS episodic retrieval and the reader backbone fixed, we replace EverMemOS's built-in user profile with our RAG-retrieved temporally-weighted profile (\S\ref{sec:chronoprofiler}), and evaluate the same three MLLMs as readers on T2 and T3 tasks (the profile does not affect T1 factual recall). As an additional T1 upper bound (Tab.~\ref{tab:oracle}), we conduct \textbf{Oracle} experiments that supply the three frontier MLLMs with only the gold-evidence sessions (identified by per-QA annotated evidence id) as context. Unless otherwise specified, all evaluations operate on a unified text representation in which media files are pre-captioned by Gemini-2.0-Flash~\cite{gemini-2.0-flash} and embedded into the dialogue transcript. Full configuration details and ChronoProfiler hyperparameters are in Appendices~\ref{app:experiment_setup} and~\ref{app:chronoprofiler}.

\noindent\textbf{Evaluation Metrics.}
We use GPT-4o-mini~\cite{gpt-4o-mini} as an LLM judge~\cite{Zheng2023JudgingLW} for \textbf{T1} task (binary correctness against the reference answer), the normalized mean of \textit{Coverage} and \textit{Accuracy} score for \textbf{T2} task, and Kendall-$\tau$ rank correlation~\cite{Kendall1938ANM} against the participant-validated ground-truth ordering for \textbf{T3} task. All metrics are scaled to $[0, 100]$; scoring rubrics and evaluation details are provided in Appendix~\ref{sec:appendix}.

\subsection{Main Results}
Tab.~\ref{tab:main_results} presents results across all evaluated MLLMs and systems, and Tab.~\ref{tab:oracle} reports the Oracle T1 upper bound.

\begin{table}[t]
\centering
\small
\setlength{\tabcolsep}{4pt}
\renewcommand{\arraystretch}{1.05}
\caption{Oracle reference on T1 Factual Memory Recall (\%). For each model the top row reports the Oracle score, the bottom row reports the absolute delta versus the same model's full-context result in Tab.~\ref{tab:main_results}.}
\label{tab:oracle}
\resizebox{\columnwidth}{!}{%
\begin{tabular}{@{}l ccccc@{}}
\toprule
\textbf{Model} & \textbf{Retr.} & \textbf{Embd.} & \textbf{Aggr.} & \textbf{M.R.} & \textbf{Avg.} \\
\midrule
\multirow{2}{*}{Gemini-3.0-Flash}
  & 87.7 & 79.6 & 83.8 & 86.9 & 85.6 \\
  & {\scriptsize($\uparrow 3.3$)} & {\scriptsize($\uparrow 1.9$)} & {\scriptsize($\uparrow 8.6$)} & {\scriptsize($\downarrow 2.4$)} & {\scriptsize($\uparrow 3.0$)} \\
\midrule
\multirow{2}{*}{GPT-4.1-mini}
  & 88.5 & 80.0 & 77.1 & 84.5 & 85.5 \\
  & {\scriptsize($\uparrow 10.5$)} & {\scriptsize($\uparrow 9.6$)} & {\scriptsize($\uparrow 33.3$)} & {\scriptsize($\uparrow 19.0$)} & {\scriptsize($\uparrow 12.9$)} \\
\midrule
\multirow{2}{*}{GPT-5.4-mini}
  & 87.8 & 80.0 & 81.9 & 83.3 & 85.4 \\
  & {\scriptsize($\uparrow 8.1$)} & {\scriptsize($\uparrow 11.1$)} & {\scriptsize($\uparrow 41.9$)} & {\scriptsize($\uparrow 26.2$)} & {\scriptsize($\uparrow 12.9$)} \\
\bottomrule
\end{tabular}}
\end{table}

\noindent\textbf{Full-context performance confirms the cognitive hierarchy.}
T3 is the uniformly hardest tier across all three frontier MLLMs (57.9--59.2\%), confirming a consistent upper bound on predictive-personalization difficulty. Above T3, the T1--T2 performance is model-dependent: Gemini-3.0-Flash follows the expected cognitive hierarchy (T1 82.6\% $>$ T2 76.4\%), whereas GPT-4.1-mini and GPT-5.4-mini both score marginally higher on T2 than T1, indicating that preference inference is not universally harder than factual recall and that long-context retrieval ability is the primary T1 differentiator. Gemini-3.0-Flash leads on both T1 (82.6\%) and T2 (76.4\%), its T1 advantage is concentrated on Multi-Record Aggregation and Multimodal Recall. On T2 and T3, the three models converge tightly, spanning 2.1 points on T2 and 1.3 points on T3, indicating that difficulty is dominated by the inherent complexity of preference reasoning and predictive personalization rather than by backbone capacity.

\begin{table}[t]
\centering
\footnotesize
\setlength{\tabcolsep}{4pt}
\renewcommand{\arraystretch}{0.95}
\caption{Effect of context modality on Gemini-3.0-Flash (\%).
  \textbf{Cap.+Dial.}: captioned dialogue with metadata; \textbf{MM}: raw visual media with
  dialogue and metadata; \textbf{Vis.+Meta.}: visual media with metadata only (no dialogue).
  Task abbreviations follow Tab.~\ref{tab:main_results}.
  Best result per row in \textbf{bold}.}
\label{tab:context_modality}
\begin{tabular}{l ccc}
\toprule
\textbf{Task} & \textbf{Cap.+Dial.} & \textbf{MM} & \textbf{Vis.+Meta.} \\
\midrule
\multicolumn{4}{l}{\textit{T1: Factual Memory Recall}} \\
\quad Retr.       & \textbf{84.4} & 82.8          & 50.7 \\
\quad Embd.       & \textbf{77.7} & 74.2          & 71.2 \\
\quad Aggr.       & 75.2          & \textbf{77.1} & 58.1 \\
\quad M.R.        & 89.3          & \textbf{90.5} & 76.2 \\
\quad \textit{Avg.} & \textbf{82.6} & 81.0        & 57.4 \\
\midrule
\multicolumn{4}{l}{\textit{T2: Persona Inference}} \\
\quad Prof.       & \textbf{74.8} & 74.5          & 61.9 \\
\quad Track       & \textbf{73.0} & 71.8          & 59.3 \\
\quad Reas.       & \textbf{81.7} & 78.1          & 66.2 \\
\quad \textit{Avg.} & \textbf{76.4} & 74.8        & 62.4 \\
\midrule
\multicolumn{4}{l}{\textit{T3: Predictive Personalization}} \\
\quad Pred.       & 62.8          & 61.3          & \textbf{64.6} \\
\quad Asst.       & 53.5          & 55.9          & \textbf{56.3} \\
\quad \textit{Avg.} & 59.2        & 59.2          & \textbf{61.4} \\
\bottomrule
\end{tabular}
\end{table}

\noindent\textbf{Memory architectures shape the compression--quality profile.}
Under the unified GPT-4.1-mini answering backbone (full-context T1/T2/T3 average $=$ 72.6/74.3/57.9\%), the three memory systems exhibit markedly different compression--quality trade-offs at ${\sim}70\times$ compression that map cleanly onto their architectural choices. Mem0~\cite{chhikara2025mem0}, which extracts flat narrative memories via LLM-driven \texttt{ADD}/\texttt{UPDATE} operations, discards precise temporal and quantitative anchors and suffers the largest T1 collapse (T1 46.8\%). MemOS~\cite{li2505memos}, with OS-style hierarchical MemCube governance, recovers more structured factual recall (T1 54.0\%) and attaches a static global preference summary to each retrieved memory; however, this uniformly-replicated, coarsely-aggregated preference string provides insufficient query-specific grounding, capping T2 at 58.8\%. EverMemOS~\cite{hu2026evermemos}, whose engram-inspired Episodic Trace and Semantic Consolidation lifecycle distils both events and preferences, tracks the full-context baseline most closely (T1 69.4\%) and attains the strongest memory-system T2 (72.0\%). Strikingly, on T3 all three systems land within about one point of the GPT-4.1-mini full-context baseline (56.8--59.2\% vs.\ 57.9\%), with Mem0 even edging ahead (59.2\%), indicating that the T3 bottleneck is representational quality rather than memory coverage.

\begin{table}[!h]
\centering
\footnotesize
\setlength{\tabcolsep}{4pt}
\renewcommand{\arraystretch}{1.05}
\caption{Profile-swap analysis on T2 and T3 (\%). \textbf{ChronoProfiler} injects our temporally-weighted profile, \textbf{Ep.+Prof.}\ EverMemOS's built-in profile; for each row, the top line is the score and the bottom line its $\Delta$ versus the same backbone's Full Context (Tab.~\ref{tab:main_results}). Best per column in \textbf{bold}.}
\label{tab:chronoprofiler}
\resizebox{\columnwidth}{!}{%
\begin{tabular}{@{}l ccc c cc c@{}}
\toprule
\multirow{2}{*}{\textbf{System}}
  & \multicolumn{4}{c}{\textbf{T2}}
  & \multicolumn{3}{c}{\textbf{T3}} \\
\cmidrule(lr){2-5}\cmidrule(lr){6-8}
 & \textbf{Prof.} & \textbf{Track} & \textbf{Reas.} & \textbf{Avg.}
 & \textbf{Pred.} & \textbf{Asst.} & \textbf{Avg.} \\
\midrule
\multicolumn{8}{l}{\textit{ChronoProfiler: Ep.+TS-Profile}} \\[1pt]
\multirow{2}{*}{\quad Gemini-3.0-Flash}
  & 76.0 & 73.1 & 82.1 & 77.0 & 61.6 & 55.4 & 59.2 \\
  & {\scriptsize($\uparrow 1.2$)} & {\scriptsize($\uparrow 0.1$)} & {\scriptsize($\uparrow 0.4$)} & {\scriptsize($\uparrow 0.6$)} & {\scriptsize($\downarrow 1.2$)} & {\scriptsize($\uparrow 1.9$)} & {\scriptsize($\pm 0.0$)} \\
\multirow{2}{*}{\quad GPT-4.1-mini}
  & 76.8 & 73.9 & \textbf{84.6} & 78.3 & \textbf{63.7} & \textbf{57.8} & \textbf{61.4} \\
  & {\scriptsize($\uparrow 2.6$)} & {\scriptsize($\uparrow 6.5$)} & {\scriptsize($\uparrow 3.6$)} & {\scriptsize($\uparrow 4.0$)} & {\scriptsize($\uparrow 3.3$)} & {\scriptsize($\uparrow 3.8$)} & {\scriptsize($\uparrow 3.5$)} \\
\multirow{2}{*}{\quad GPT-5.4-mini}
  & \textbf{77.9} & \textbf{78.0} & 84.0 & \textbf{79.8} & \textbf{63.7} & 56.3 & 60.8 \\
  & {\scriptsize($\uparrow 2.0$)} & {\scriptsize($\uparrow 9.1$)} & {\scriptsize($\uparrow 2.7$)} & {\scriptsize($\uparrow 4.3$)} & {\scriptsize($\uparrow 4.5$)} & {\scriptsize($\downarrow 2.4$)} & {\scriptsize($\uparrow 1.8$)} \\
\midrule
\multicolumn{8}{l}{\textit{EverMemOS: Ep.+Prof.}} \\[1pt]
\multirow{2}{*}{\quad GPT-4.1-mini}
  & 71.2 & 67.0 & 77.8 & 72.0 & 63.1 & 53.5 & 59.3 \\
  & {\scriptsize($\downarrow 3.0$)} & {\scriptsize($\downarrow 0.4$)} & {\scriptsize($\downarrow 3.2$)} & {\scriptsize($\downarrow 2.3$)} & {\scriptsize($\uparrow 2.7$)} & {\scriptsize($\downarrow 0.5$)} & {\scriptsize($\uparrow 1.4$)} \\
\bottomrule
\end{tabular}}
\end{table}

\noindent\textbf{Oracle exposes a retrieval bottleneck on T1.}
All three MLLMs converge to a similar Oracle ceiling (85.4--85.6\% T1 Avg.), but their full-context gaps diverge sharply: Gemini trails its Oracle by only $+3.0\%$, whereas GPT-4.1-mini and GPT-5.4-mini both lag far more ($+12.9\%$ each). This shows that Gemini's T1 advantage stems primarily from robustness to distractor evidence in long context, and memory systems implicitly inherit the task of approximating an oracle retriever---which is why EverMemOS can close most of the GPT-4.1-mini full-context gap with only 1.7k tokens.

\noindent\textbf{T3 reveals a universal ceiling unaddressed by context length or retrieval.}
Across every system in Tab.~\ref{tab:main_results}, T3 remains the hardest level, with the highest T3 Avg.\ capped at 59.2\% and the full spread under 3 points (56.8--59.2\%). Unlike T1, T3 has no evidence-bounded upper bound: the bottleneck is not retrieval or compression but the fine-grained discrimination of preference signals differentially influence decisions across contexts---recognizing that a user holds a preference is necessary but insufficient; the model must further judge which preferences dominate, when they trade off, and how their relative weights shift with situational context.

\subsection{What Each Modality Contributes}
\label{sec:modality}
To probe the effect of context modality, we compare three input settings: captioned dialogue with spatio-temporal metadata (our default), raw multimodal media with dialogue and metadata (MM), and visual media with metadata but no dialogue (Vis.+Meta.). Because raw multimodal input far exceeds the captioned context length, we run this ablation on Gemini-3.0-Flash, the reader that natively ingests long-context raw multimodal input (Tab.~\ref{tab:context_modality}).

\noindent\textbf{Captions are a faithful proxy for raw pixels on factual recall.}
With dialogue held fixed, captioned dialogue and MM differ only marginally---captions slightly help text-anchored subtasks (episodic and embedded retrieval, all of T2), while raw pixels slightly help the two subtasks needing quantitative or visual reasoning (Multi-Record Aggregation and Multimodal Recall). The overall T1 gap is just $1.6$ points (82.6 vs.\ 81.0), showing that high-quality MLLM captions preserve most factual content and validating captioned dialogue as the unified input for fair comparison across memory systems.

\noindent\textbf{Language-mediated context dominates recall and inference tasks.}
Holding the visual input fixed and removing only the dialogue stream (MM $\to$ Vis.+Meta.) collapses T1 by $23.6$ points (81.0$\to$57.4) and T2 by $12.4$ (74.8$\to$62.4), sharpest on Episodic Event Retrieval ($-32.1$). 
The drop is this steep because most of ReaLMem's factual grounding, together with all of its first-person subjective annotations resides in the textual stream: raw frames depict scenes but not the identities, labels, and personal context that T1 recall and T2 inference require.

\noindent\textbf{Vision alone is the strongest signal for prediction.}
T3 breaks the monotone pattern: the dialogue-free setting (Vis.+Meta.) attains the highest T3 Avg. ($61.4$ vs.\ $59.2$) and leads on both subtasks. 
We attribute this gap to the nature of the signal: the visual stream itself reflects a user's preferences and the relative composition of their life. A rich, multi-year dialogue, by contrast, injects mutually conflicting and uneven signals that the model cannot reliably arbitrate, whereas the sparser visual-and-metadata view falls back on these broad, stable patterns of activity and place. Since predictive personalization is a ranking task that rewards a holistic read of a user's habitual preferences over fine-grained arbitration of scattered textual cues, this sparser view wins.

\subsection{ChronoProfiler Analysis}
\label{sec:chronoprofiler_analysis}

We probe ChronoProfiler with two controlled experiments, both evaluating the three MLLMs on T2 and T3 with deltas reported against the same backbone's Full Context. First, a \emph{profile swap} (Tab.~\ref{tab:chronoprofiler}) holds EverMemOS episodic retrieval and replaces EverMemOS's built-in profile with our temporally-weighted profile; comparing against Full Context and against \textbf{EverMemOS: Ep.+Prof.} (the identical pipeline with EverMemOS's own profile) isolates the profile representation from raw context length and from the quality of a profile. Second, a \emph{TS ablation} (Tab.~\ref{tab:tw_ablation}) fixes the retrieval depth (top-15 items retrieval per category) and compares the profile with and without temporal stability weighting, isolating the contribution of TS. 
We additionally study how the retrieval depth affects performance, the configuration and full depth--accuracy analysis are deferred to Appendix~\ref{app:retrieval_depth}, with all main experiments fixing the depth at top-15. Together, these analyses separate the factors behind the gains: context length, profile representation, TS weighting, and retrieval depth.

\noindent\textbf{ChronoProfiler matches or beats Full Context on both T2 and T3.}
At only ${\sim}6$k prompt tokens, roughly $20\times$ fewer than the Full Context, ChronoProfiler exceeds full context on T2 Avg. for all three readers ($\Delta=+0.6/+4.0/+4.3$) and matches or exceeds it on T3 Avg. ($\Delta=\pm0.0/+3.5/+1.8$). The strongest configuration is GPT-4.1-mini, the only reader to improve on \emph{every} subtask, posting the table-best T3 Avg. (61.4\%, $\Delta=+3.5$) and the best Context-Aware Assistance (57.8\%). This confirms that the context sufficient for preference-grounded inference is far smaller than the raw history, provided it captures the right temporal structure. 

\noindent\textbf{The temporally-weighted profile dominates EverMemOS's built-in profile.}
Holding episodes and the GPT-4.1-mini backbone fixed, swapping in our profile improves \emph{every} T2 metric (T2 Avg. $+6.3$; Persona Profiling $+5.6$, Longitudinal Tracking $+6.9$, Behavioral Reasoning $+6.8$) and both T3 metrics (T3 Avg. $+2.1$; Context-Aware Assistance $+4.3$). Tellingly, EverMemOS's own profile \emph{degrades} Full Context on T2 ($\Delta=-2.3$) while ours \emph{lifts} it ($\Delta=+4.0$), pinpointing the profile representation---not the episodic memory---as the active ingredient behind the T2 and T3 gains.

\noindent\textbf{Temporal stability is the source of the T2 gains.}
Adding TS lifts T2 Avg. for all three readers ($+2.7/+2.0/+0.3$) and raises Persona Profiling across the board ($+3.0/+2.6/+1.1$), directly evidencing TS as a time-aware prior for consolidating stable and evolving preferences. On T3 the effect is comparable or slightly lower (T3 Avg $+0.6/-0.4/-1.2$): TS consistently helps Context-Aware Prediction for both GPT readers ($+2.4/+0.9$) but uniformly trades off a small amount of Context-Aware Assistance (${\approx}-2.4$), which we attribute to TS sharpening recurring preferences at the expense of the broader, occasion-specific cues that open-ended recommendation can exploit. We therefore scope the TS claim to persona profiling and longitudinal tracking, and regard selective episodic retrieval as a complementary route to recovering the Assistance gap.

\begin{table}[t]
\centering
\footnotesize
\setlength{\tabcolsep}{4pt}
\renewcommand{\arraystretch}{1.05}
\caption{Temporal-stability (TS) ablation on T2 and T3 (\%). Rows compare the profile without (\emph{w/o TS}) and with (\textbf{+\,TS}) TS weighting; $\Delta$ is the gain from adding TS.}
\label{tab:tw_ablation}
\resizebox{\columnwidth}{!}{%
\begin{tabular}{@{}l ccc c cc c@{}}
\toprule
\multirow{2}{*}{\textbf{Setting}}
  & \multicolumn{4}{c}{\textbf{T2}}
  & \multicolumn{3}{c}{\textbf{T3}} \\
\cmidrule(lr){2-5}\cmidrule(lr){6-8}
 & \textbf{Prof.} & \textbf{Track} & \textbf{Reas.} & \textbf{Avg.}
 & \textbf{Pred.} & \textbf{Asst.} & \textbf{Avg.} \\
\midrule
\multicolumn{8}{l}{\textit{GPT-4.1-mini}} \\[1pt]
\quad w/o TS         & 73.8 & 72.0 & 81.7 & 75.6 & 61.3 & 60.1 & 60.8 \\
\quad +\,TS (Ours)   & 76.8 & 73.9 & 84.6 & 78.3 & 63.7 & 57.8 & 61.4 \\
\quad \textit{$\Delta$} & {\scriptsize($\uparrow 3.0$)} & {\scriptsize($\uparrow 1.9$)} & {\scriptsize($\uparrow 2.9$)} & {\scriptsize($\uparrow 2.7$)} & {\scriptsize($\uparrow 2.4$)} & {\scriptsize($\downarrow 2.3$)} & {\scriptsize($\uparrow 0.6$)} \\
\midrule
\multicolumn{8}{l}{\textit{GPT-5.4-mini}} \\[1pt]
\quad w/o TS         & 75.3 & 76.4 & 82.4 & 77.8 & 62.8 & 58.7 & 61.2 \\
\quad +\,TS (Ours)   & 77.9 & 78.0 & 84.0 & 79.8 & 63.7 & 56.3 & 60.8 \\
\quad \textit{$\Delta$} & {\scriptsize($\uparrow 2.6$)} & {\scriptsize($\uparrow 1.6$)} & {\scriptsize($\uparrow 1.6$)} & {\scriptsize($\uparrow 2.0$)} & {\scriptsize($\uparrow 0.9$)} & {\scriptsize($\downarrow 2.4$)} & {\scriptsize($\downarrow 0.4$)} \\
\midrule
\multicolumn{8}{l}{\textit{Gemini-3.0-Flash}} \\[1pt]
\quad w/o TS         & 74.9 & 73.6 & 81.9 & 76.7 & 62.2 & 57.8 & 60.4 \\
\quad +\,TS (Ours)   & 76.0 & 73.1 & 82.1 & 77.0 & 61.6 & 55.4 & 59.2 \\
\quad \textit{$\Delta$} & {\scriptsize($\uparrow 1.1$)} & {\scriptsize($\downarrow 0.5$)} & {\scriptsize($\uparrow 0.2$)} & {\scriptsize($\uparrow 0.3$)} & {\scriptsize($\downarrow 0.6$)} & {\scriptsize($\downarrow 2.4$)} & {\scriptsize($\downarrow 1.2$)} \\
\bottomrule
\end{tabular}}
\end{table}

\section{Conclusion and Limitations}
We introduced ReaLMem, the first multimodal long-term memory benchmark grounded in authentic personal visual archives and first-person subjective annotations, organized as a Bloom-inspired three-tier hierarchy from factual recall through persona inference to predictive personalization. To counter the preference-flattening of existing memory systems, we proposed ChronoProfiler, a memory-consolidation-inspired module that scores each attribute by its temporal stability and injects it as a salience prior. Our experiments identify predictive personalization as the central bottleneck and show that a compact temporally-weighted profile can rival full context, establishing temporally-weighted preference selection as a promising direction for real-world personal-memory modeling.

ReaLMem deliberately trades breadth for depth: authentic, long-horizon first-person archives are bounded by acquisition cost and strict privacy, so the participant pool is small, and ChronoProfiler's decay constants follow cognitive principles rather than population-scale tuning. As larger and more diverse personal-memory corpora emerge, broader-coverage benchmarks and adaptive, user-specific temporal-stability parameterizations---learned from each user's own engagement rhythm---are natural next steps.

{
    \small
    \bibliographystyle{ieeenat_fullname}
    \bibliography{main}
}

\maketitlesupplementary
\appendix
\section{Ethical Considerations}
\label{app:ethics}
The design of ReaLMem, including its data collection, processing, and annotation, underwent ethical review and was approved by the ethical review board of the authors' organization. Prior to participation, each contributor was fully informed of the data-collection procedure, the categories of personal content involved, and the intended research use, and signed an informed-consent agreement. Beyond consent, every record passes through the anonymization pipeline of Appendix~\ref{app:anonymization} (face blurring, PII pseudonymization, and metadata sanitization) before any downstream use. To guard against potential negative applications, ReaLMem will be distributed for research purposes only and released exclusively to users who sign a data license agreement governing acceptable use.

\section{Dataset Anonymization}
\label{app:anonymization}
To protect participant privacy before any processing, we detect faces with RetinaFace~\cite{Deng2020CVPR} at a detection confidence threshold of $0.5$, and apply Gaussian blur to every detected face region. In parallel, personally identifiable information in text---names, precise locations, and contact details---is pseudonymized with stable placeholders (\eg, ``Friend~A'', ``[Restaurant~B]'') that preserve referential consistency for analysis while preventing re-identification. Finally, we sanitize the raw metadata so that the released dataset contains only anonymized, privacy-safe records.

\section{Experimental Setup Details}
\label{app:experiment_setup}

\paragraph{Memory system configurations.}
Mem0~\cite{chhikara2025mem0}, MemOS~\cite{li2505memos}, and EverMemOS~\cite{hu2026evermemos} are run under each system's default configuration during the memory construction stage; we replace their answering backbone with a unified GPT-4.1-mini to isolate the contribution of the memory architecture from the underlying MLLM. The resulting per-query inference context ranges from 1.0k (Mem0) to 1.7k (EverMemOS) tokens.

\paragraph{Oracle reference setup.}
The Oracle setting supplies each frontier MLLM with only the gold-evidence sessions for the current question, identified by the per-QA evidence annotation. Oracle is restricted to T1 because T2 and T3 require holistic synthesis across the full participant history rather than retrieval of a localized evidence span.

\section{Evaluation Details}
\label{sec:appendix}
\paragraph{T1 Factual Memory Recall --- judge prompt.}
T1 responses are evaluated by GPT-4o-mini as \textsc{Correct} or \textsc{Wrong}. The judge is provided with the question, the reference answer, and the model's response and applies four criteria: (1) \textit{Conceptual Match} --- paraphrase and non-essential omissions are accepted as long as core entities are preserved; (2) \textit{Time \& Dates} --- relative and format-variant references are accepted if they resolve to the same calendar date (year, month, day); time-of-day differences are ignored; (3) \textit{Numbers \& Counts} --- summaries requiring exact counts must match the reference; (4) \textit{No Contradictory Hallucinations} --- a response that includes correct information alongside confidently stated contradictory fabrications is marked \textsc{Wrong}. The T1 score for an instance is 1 (\textsc{Correct}) or 0 (\textsc{Wrong}); subtask scores are the mean over QA pairs.
\begin{table*}[!t]
\centering
\footnotesize
\renewcommand{\arraystretch}{1.15}
\setlength{\tabcolsep}{4pt}
\begin{tabular}{@{} p{0.12\linewidth} p{0.20\linewidth} p{0.34\linewidth} p{0.30\linewidth} @{}}
\toprule
\textbf{Category} & \textbf{Sub-task (\#QA)} & \textbf{Definition} & \textbf{Example Query} \\
\midrule

\multirow{6}{=}{\textbf{1. Factual Memory Recall}}
 & \textbf{1.1 Episodic Event Retrieval} \textit{(789)} & Retrieve specific details (when, where, who, what) of a past event. & \textit{``When was the last time I dined at [Restaurant B] with [Friend A]?''} \\
 \cmidrule(l){2-4}
 & \textbf{1.2 Embedded Content Retrieval} \textit{(260)} & Retrieve information embedded in visual media (\eg, text in screenshots). & \textit{``What is the flight number in the booking confirmation screenshot I saved?''} \\
 \cmidrule(l){2-4}
 & \textbf{1.3 Multi-Record Aggregation} \textit{(105)} & Aggregate, count, or compare information across multiple memory records. & \textit{``How many times did I visit the gym this year?''} \\
 \cmidrule(l){2-4}
 & \textbf{1.4 Multimodal Recall} \textit{(84)}& Retrieve relevant memories grounded in visual content. & \textit{``[Photo] When was the last time I ate at this restaurant?''} \\
\midrule

\multirow{5}{=}{\textbf{2. Persona Inference}}
 & \textbf{2.1 Persona Profiling} \textit{(85)} & Profile and characterize stable user preferences, habits, and recurring behavioral patterns. & \textit{``What is my usual morning routine?''} \\
 \cmidrule(l){2-4}
 & \textbf{2.2 Longitudinal Tracking} \textit{(61)} & Detect changes or evolution in preferences and behaviors over time. & \textit{``Do I prefer indoor or outdoor sports now compared to two years ago?''} \\
 \cmidrule(l){2-4}
 & \textbf{2.3 Behavioral Reasoning} \textit{(63)} & Infer the underlying reasons or motivations behind observed behavioral changes. & \textit{``Why did I seemingly stop going to [Gym A]?''} \\
\midrule

\multirow{3}{=}{\textbf{3. Predictive Personalization}}
 & \textbf{3.1 Context-Aware Prediction} \textit{(111)} & Predict user preference or choice in a new scenario conditioned on personal history. & \textit{``For a winter trip, would I enjoy [Location A] or [Location B] more?''} \\
 \cmidrule(l){2-4}
 & \textbf{3.2 Context-Aware Assistance} \textit{(71)} & Provide actionable, personalized recommendations grounded in the user's personal history. & \textit{``Suggest a suitable gift for [Friend A] based on our shared memories.''} \\
\bottomrule
\end{tabular}
\caption{Taxonomy of the ReaLMem benchmark. Nine subtasks are organized into three cognitive categories spanning factual memory recall, persona inference, and predictive personalization. QA pair counts per subtask are shown in parentheses.}
\label{tab:task_overview}
\end{table*}

\paragraph{T2 Persona Inference --- scoring rubric.}
Each T2 instance provides a question, a reference answer, and participant-validated key supporting points. GPT-4o-mini assigns two independent integer scores on a 1--5 scale:
\begin{itemize}[noitemsep,topsep=2pt]
  \item \textbf{Coverage} (1--5): degree to which the response addresses the key supporting points --- 5 covers all comprehensively, 3 covers roughly half, 1 misses the point entirely.
  \item \textbf{Accuracy} (1--5): logical soundness and factual consistency with the reference --- 5 is fully accurate with no hallucinations, 3 has minor misinterpretations, 1 is completely fabricated or contradictory.
\end{itemize}
The reported T2 score for an instance is $(\text{Coverage} + \text{Accuracy}) \times 10$, giving a range of $[20, 100]$.

\paragraph{T3 Predictive Personalization --- ranking metric.}
Each T3 instance presents a set of candidate options and requires the model to produce a ranking. Evaluation uses Kendall-$\tau$ rank correlation between the predicted and participant-validated ground-truth ranking:
\[
  \tau = \frac{(\text{\# concordant pairs}) - (\text{\# discordant pairs})}{\binom{n}{2}},
\]
which is then affinely rescaled to $[0, 100]$ via $\text{score} = 50\cdot(\tau + 1)$, so that random ranking yields 50 and a perfect reversal yields 0.

\paragraph{Aggregation.}
Within each subtask, scores are averaged uniformly over QA pairs. Each cognitive level (T1/T2/T3) reports the QA-count-weighted mean over its constituent subtasks.

\section{QA Generation Details}
\label{app:qa_generation}

We provide the full construction procedure for the 1{,}629 QA pairs in \textsc{ReaLMem}, summarised in §3.2 of the main text.

\paragraph{T1 Factual Memory Recall.}
T1 QA pairs are generated from objectively annotated memory events. Each instance is formulated as an open-ended question accompanied by a reference answer and explicit grounding evidence (data IDs). To ensure linguistic naturalness and evaluation validity, all generated QA pairs undergo multi-stage human quality control by trained annotators, who address unnatural phrasing, insufficient contextual constraints, improper temporal expressions, and low-information or ambiguous answers, yielding high-quality, well-specified queries.

\paragraph{T2 Persona Inference.}
T2 QA generation leverages subjective annotations capturing user preferences and personal contexts. Because these tasks are inherently subjective, each QA pair consists of a question, a reference answer, and a set of key supporting points used for graded evaluation, rather than a single rigid ground truth. Crucially, these instances are verified and refined by the original data contributors, who assess the completeness, correctness, and relevance of the answer against their own lived experiences, ensuring that subjective reasoning tasks remain both reliable and faithful to real user perspectives.

\paragraph{T3 Predictive Personalization.}
For T3, we move beyond conventional multiple-choice QA and adopt a ranking-based evaluation paradigm that better reflects real-world decision-making. Instead of collapsing preference modeling and application into binary or single-choice outputs, each query presents multiple candidate options constructed from different combinations of user preferences, and the task requires models to produce a ranked ordering aligned with the user's actual preference distribution. The ground-truth rankings are directly provided and validated by the participants, capturing the nuanced, multi-dimensional, and context-dependent nature of human decision-making. This design avoids oversimplified preference matching and instead evaluates whether models can integrate heterogeneous signals and perform flexible, personalized reasoning.

After manual screening and revision, the pipeline yields a total of 1{,}629 QA pairs that are grounded, diverse, and reliable, while fully leveraging the advantages of real-world data and human-in-the-loop validation.

\section{ChronoProfiler Implementation Details}
\label{app:chronoprofiler}

\paragraph{Update operations.}
During incremental profile construction (Algorithm~\ref{alg:construction}), each newly extracted candidate is compared against existing entries in the same semantic category and resolved by one of four operations: \textsc{New} inserts a previously unseen attribute as a fresh entry; \textsc{Duplicate} keeps the existing entry unchanged and only appends the current session to its evidence list; \textsc{Update} overwrites the entry with refined or more specific content while extending its evidence list; \textsc{Conflict} (a contradiction or change relative to an existing entry, \eg ``used to like coffee'' vs.\ ``switched to tea'') does not overwrite the prior entry; instead it instantiates the new attribute as a separate entry that links back to the conflicting one via a related reference, so that the outdated and current signals are both retained as independently timestamped items, their competition left to be arbitrated downstream by the Temporal Stability score rather than by destructive overwriting.

\paragraph{Hyperparameters.}
The decay constants $\tau_s$, $\tau_d$, and $\tau_r$ control how quickly each lifecycle feature saturates. We set $\tau_s=3$, so that support saturates around three independent mentions; $\tau_d=150$~days, so duration grows substantially over roughly five months of consistent expression; and $\tau_r=60$~days, giving recency a half-life on the order of two months. Unlike the feature timescales, the combination weights $\mathbf{w}=(w_s, w_d, w_r)$ are \emph{not} fixed hyperparameters: they are recomputed per user by the entropy weight method over that user's profile-item feature matrix, so a feature that fails to discriminate among a user's items is automatically down-weighted. The resulting entropy-weighted sum is then min--max normalized within each user to $[0,1]$. The reference time $T_{\mathrm{ref}}$ is set to a common evaluation timestamp shared across users (here we set it to 2026-01-12). At inference, profile items are retrieved per semantic category by dense query similarity (computed with OpenAI's text-embedding-3-small~\cite{text-embedding-3}) and truncated to the top-15 per category (depth sensitivity is analyzed in Appendix~\ref{app:retrieval_depth}).

\paragraph{Profile-swap evaluation protocol.}
For the ChronoProfiler results in Tab. 5 and 6 of the main paper, we hold EverMemOS's episodic retrieval and the reader backbone fixed and substitute only the injected user profile, replacing EverMemOS's built-in profile with our temporally-weighted profile. This yields a per-query context of ${\sim}6$k tokens and forms a controlled contrast against the \textbf{EverMemOS: Ep.+Prof.} baseline (identical episodes and backbone, EverMemOS's own profile), so that any difference isolates the effect of the profile representation and its temporal stability weighting from both episodic retrieval and raw long-context access.

\begin{algorithm}[h]
\caption{Incremental Profile Construction}
\label{alg:construction}
\begin{algorithmic}[1]
\REQUIRE Session sequence $\mathcal{D} = \{d_1, \dots, d_N\}$ in chronological order.
\ENSURE Profile $\mathcal{P}$ with timestamped evidence.
\STATE $\mathcal{P} \leftarrow$ Initialize empty profile with 5 semantic categories
\FOR{each session $d_t \in \mathcal{D}$}
    \STATE $\mathcal{E}_t \leftarrow \text{LLM.Extract}(d_t)$
    \FOR{each candidate $e \in \mathcal{E}_t$}
        \STATE $relation \leftarrow \text{LLM.Compare}(e, \mathcal{P}[e.category])$
        \IF{$relation ==$ \textsc{New}}
            \STATE $\mathcal{P}[e.category]\text{.Append}(e, \text{evidence}=\{d_t\})$
        \ELSIF{$relation ==$ \textsc{Duplicate}}
            \STATE $match\text{.evidence.Add}(d_t)$
        \ELSIF{$relation ==$ \textsc{Update}}
            \STATE $match\text{.Update}(e)$; $match\text{.evidence.Add}(d_t)$
        \ELSIF{$relation ==$ \textsc{Conflict}}
            \STATE $\mathcal{P}[e.category]\text{.Append}(e, \text{evidence}=\{d_t\}, \text{related\_id}=match.id)$
        \ENDIF
    \ENDFOR
    \STATE $\mathcal{P}\text{.SaveCheckpoint}()$
\ENDFOR
\RETURN $\mathcal{P}$
\end{algorithmic}
\end{algorithm}

\begin{algorithm}[h]
\caption{Temporal Stability Scoring}
\label{alg:scoring}
\begin{algorithmic}[1]
\REQUIRE Profile $\mathcal{P}$, Reference time $T_{\mathrm{ref}}$, Decay constants $\tau_s=3, \tau_d=150, \tau_r=60$.
\ENSURE Profile $\mathcal{P}$ with TS score for each item.
\FOR{each item $p \in \mathcal{P}$}
    \STATE $\mathcal{S}_p \leftarrow |\text{Unique}(p\text{.mentioned\_times})|$
    \STATE $t_{\mathrm{first}}, t_{\mathrm{last}} \leftarrow \text{MinMax}(p\text{.mentioned\_times})$
    \STATE $s_{\mathrm{sup}} \leftarrow 1 - \exp(-\mathcal{S}_p / \tau_s)$
    \STATE $s_{\mathrm{dur}} \leftarrow 1 - \exp(-(t_{\mathrm{last}} - t_{\mathrm{first}}) / \tau_d)$
    \STATE $s_{\mathrm{rec}} \leftarrow \exp(-(T_{\mathrm{ref}} - t_{\mathrm{last}}) / \tau_r)$
\ENDFOR
\STATE $\mathbf{w} \leftarrow \text{EntropyWeights}(\{(s_{\mathrm{sup}}, s_{\mathrm{dur}}, s_{\mathrm{rec}})_p\}_{p\in\mathcal{P}})$ \COMMENT{per-user}
\FOR{each item $p \in \mathcal{P}$}
    \STATE $z_p \leftarrow w_s \cdot s_{\mathrm{sup}} + w_d \cdot s_{\mathrm{dur}} + w_r \cdot s_{\mathrm{rec}}$
\ENDFOR
\STATE $p\text{.TS} \leftarrow (z_p - \min_q z_q) / (\max_q z_q - \min_q z_q),\ \forall p$ \COMMENT{within-user min--max}
\RETURN $\mathcal{P}$
\end{algorithmic}
\end{algorithm}

\begin{figure*}[t!]
  \centering
  \includegraphics[width=0.9\linewidth]{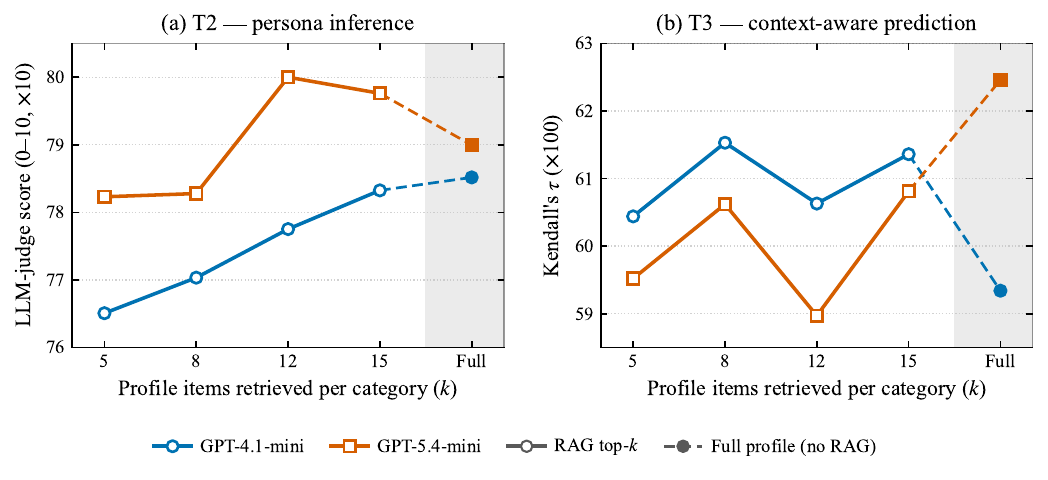}
  \caption{ChronoProfiler retrieval-depth sensitivity for the two GPT readers.
    Each panel plots the temporally-weighted profile under per-category RAG as a
    function of retrieval depth $k$ (solid, hollow markers), with the full no-RAG
    profile shown as the right-most \emph{Full} tick (filled markers, shaded).
    (a) T2 persona understanding (LLM-judge $0$--$10$, $\times10$);
    (b) T3 context-aware reasoning (Kendall's $\tau$, $\times100$).}
  \label{fig:chronoprofiler_k}
\end{figure*}

\section{ChronoProfiler Retrieval-Depth Sensitivity}
\label{app:retrieval_depth}

 We fix the per-category retrieval depth at $k=15$ and report the results in Tab. 5 and 6 of the main paper. Figure~\ref{fig:chronoprofiler_k} traces how our temporally-weighted profile behaves under per-category RAG as $k$ varies, for the two GPT readers, with the full no-RAG profile (${\sim}16$k tokens) shown as the right-most categorical tick for reference.

\paragraph{T2 saturates well before the full budget.}
Persona understanding (T2) rises smoothly with $k$ and is essentially flat by $k=15$, where both readers sit within ${\sim}0.8$ points of their full-profile score (GPT-4.1-mini: $78.3$ vs.\ $78.5$; GPT-5.4-mini: $79.8$ vs.\ $79.0$) while using only ${\sim}6$k of the ${\sim}16$k full-profile tokens (${\approx}40\%$). Retrieving more profile items therefore buys little on persona inference, which motivates our top-15 default.

\paragraph{T3 is reader-dependent.}
Context-aware prediction (T3) does not share a single optimum. GPT-4.1-mini peaks at the RAG operating point ($k=15$, $61.4$) and \emph{declines} when given the full profile ($59.3$), consistent with the dialogue-dilution effect we observe in the modality ablation (\S5.3): a longer, noisier preference record introduces conflicting signals that this reader cannot arbitrate. GPT-5.4-mini, in contrast, continues to benefit from the full profile ($62.5$ vs.\ $60.8$ at $k=15$). Because the compact top-15 profile is at or near the best T3 setting for GPT-4.1-mini and within ${\sim}1.7$ points for GPT-5.4-mini, at a fraction of the token cost, we adopt $k=15$ as the single default across readers.

\end{document}